\documentclass[letterpaper, 10 pt, conference]{support/ieeeconf}
\IEEEoverridecommandlockouts
\usepackage{cite}
\usepackage{amsmath,amssymb,amsfonts}
\usepackage{algorithmic}
\usepackage{graphicx}
\usepackage{textcomp}
\usepackage{multirow}
\usepackage{etoolbox}
\usepackage{url}
\usepackage{booktabs}
\usepackage[ruled,linesnumbered]{algorithm2e}

\makeatletter
\makeatletter
\def\endthebibliography{%
	\def\@noitemerr{\@latex@warning{Empty `thebibliography' environment}}%
	\endlist
}

\patchcmd{\@makecaption}
  {\scshape}
  {}
  {}
  {}
\makeatletter
\patchcmd{\@makecaption}
  {\\}
  {.\ }
  {}
  {}
\makeatother
\def\tablename{Table}   

\def\BibTeX{{\rm B\kern-.05em{\sc i\kern-.025em b}\kern-.08em
    T\kern-.1667em\lower.7ex\hbox{E}\kern-.125emX}}
\begin{document}
\title{Decentralized Planning for Car-Like Robotic Swarm in Cluttered Environments\\
    \thanks{$^{1}$State Key Laboratory of Industrial Control Technology, Zhejiang University, Hangzhou 310027, China.(\textit{Corresponding author: Fei Gao})}
    \thanks{$^{2}$Huzhou Institute of Zhejiang University, Huzhou 313000, China.}
    \thanks{E-mail: {\tt\small \{changjiama and fgaoaa\}@zju.edu.cn}}
}
\author{Changjia Ma$^{1,2}$, Zhichao Han$^{1,2}$, Tingrui Zhang$^{1,2}$,\\ Jingping Wang$^{1,2}$, Long Xu$^{1,2}$, Chengyang Li$^{2}$, Chao Xu$^{1,2}$ and Fei Gao$^{1,2}$}

\maketitle

\begin{abstract}
Robot swarm is a hot spot in robotic research community. 
In this paper, we propose a decentralized framework for car-like robotic swarm which is capable of real-time planning in cluttered environments. 
In this system, path finding is guided by environmental topology information to avoid frequent topological change, and search-based speed planning is leveraged to escape from infeasible initial value's local minima. 
Then spatial-temporal optimization is employed to generate a safe, smooth and dynamically feasible trajectory. 
During optimization, the trajectory is discretized by fixed time steps. 
Penalty is imposed on the signed distance between agents to realize collision avoidance, and differential flatness cooperated with limitation on front steer angle satisfies the non-holonomic constraints. 
With trajectories broadcast to the wireless network, agents are able to check and prevent potential collisions. We validate the robustness of our system in simulation and real-world experiments. 
Code will be released as open-source packages. \looseness=-1
\end{abstract}

\section{Introduction}
Benefiting from the flexibility and stability, multi-robot systems can significantly accelerate task completion. 
Nowadays, car-like robotic swarms are widely applied in real life, such as autonomous trucks in ports, automatic guided vehicles in logistics warehouses, search and rescue in hazardous scenarios, and exploration in unknown environments. 
Most of these systems rely on centralized planning frameworks.
However, with the robot quantity and map complexity increasing, centralized methods suffer from high computation burden, hence they lack the adaptability to dynamic environments, restricting the practical usage of car-like robotic swarms.
For decentralized methods, the computation burden is separated by each agent. 
But limited by the performance of onboard processors, trajectory quality may not be guaranteed.
To fill this gap, we propose a decentralized planning framework for car-like robotic swarm which is capable of real-time planning in obstacle-dense environments.

\begin{figure}[htp]
	\centering
	\includegraphics[width=9cm]{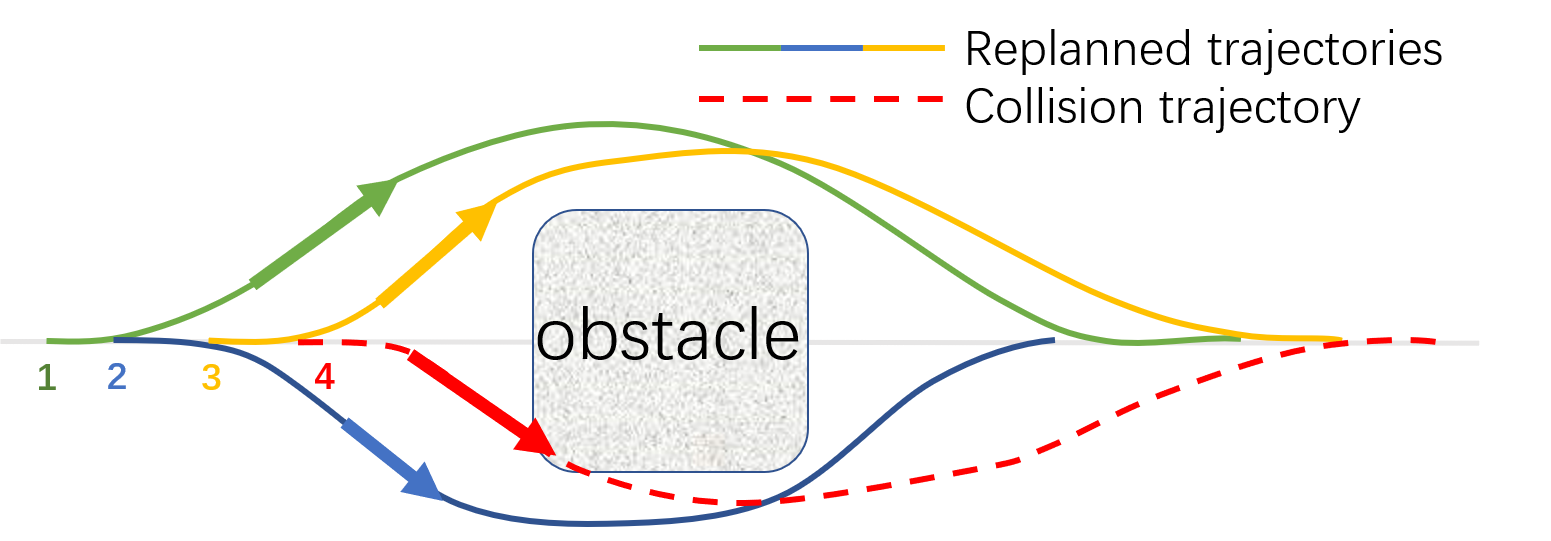}
	\caption{The green, blue, yellow and red curves are successively replanned trajectories. The agent is near the obstacle, but frequent replans lead to the topological non-homotopy, which makes the red curve intersect with the obstacle.}
	\label{fig: frequent replan collision}
	\vspace{-0.5cm}
\end{figure}

Different from robots with holonomic kinematics such as quadrotors, car-like robots with non-holonomic and shape constraints bring more challenges for online planning in cluttered environments.
During navigation, it is necessary to conduct frequent replans to avoid collisions with static and dynamic obstacles. 
Due to the non-holonomic kinematics, adjusting to rapid change in trajectories' topological structure is more difficult for car-like robots, which results in a higher rate of colliding with static obstacles, especially when the agent is close to the obstacle, as shown in Fig.\ref{fig: frequent replan collision}. 
To overcome this difficulty, we introduce topology-guided path planning to guarantee trajectory consistency, which means that the replanned path remains topology homotopy with the last planned one.

\begin{figure}[t]
    \vspace{0.2cm}
    \centering
    \setlength{\abovecaptionskip}{-0.00cm}
    \includegraphics[width=8.5cm]{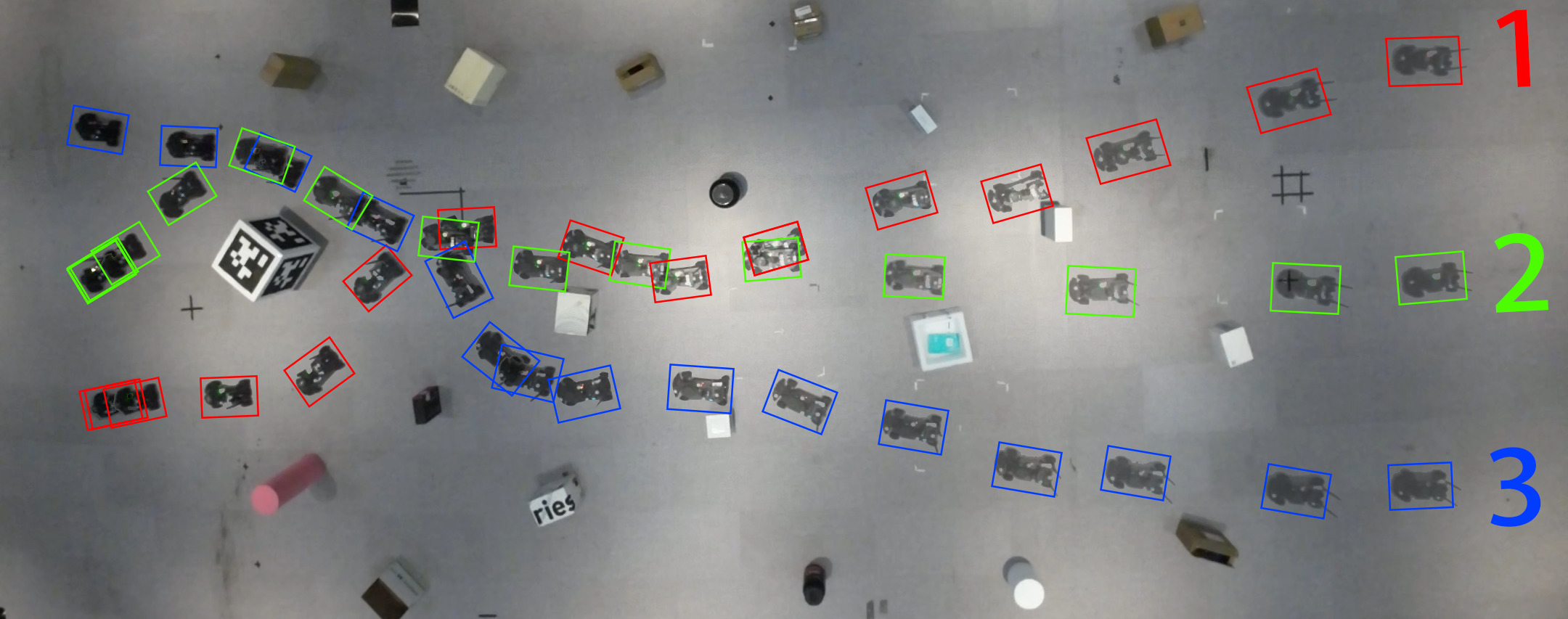}
    \caption{Three car-like robots traverse an unexplored environment.}
    \label{fig:real world experiments}
    \vspace{-0.35cm}
\end{figure}
\begin{figure}[t]
    \centering
    \setlength{\abovecaptionskip}{-0.05cm}
    \includegraphics[width=8.5cm]{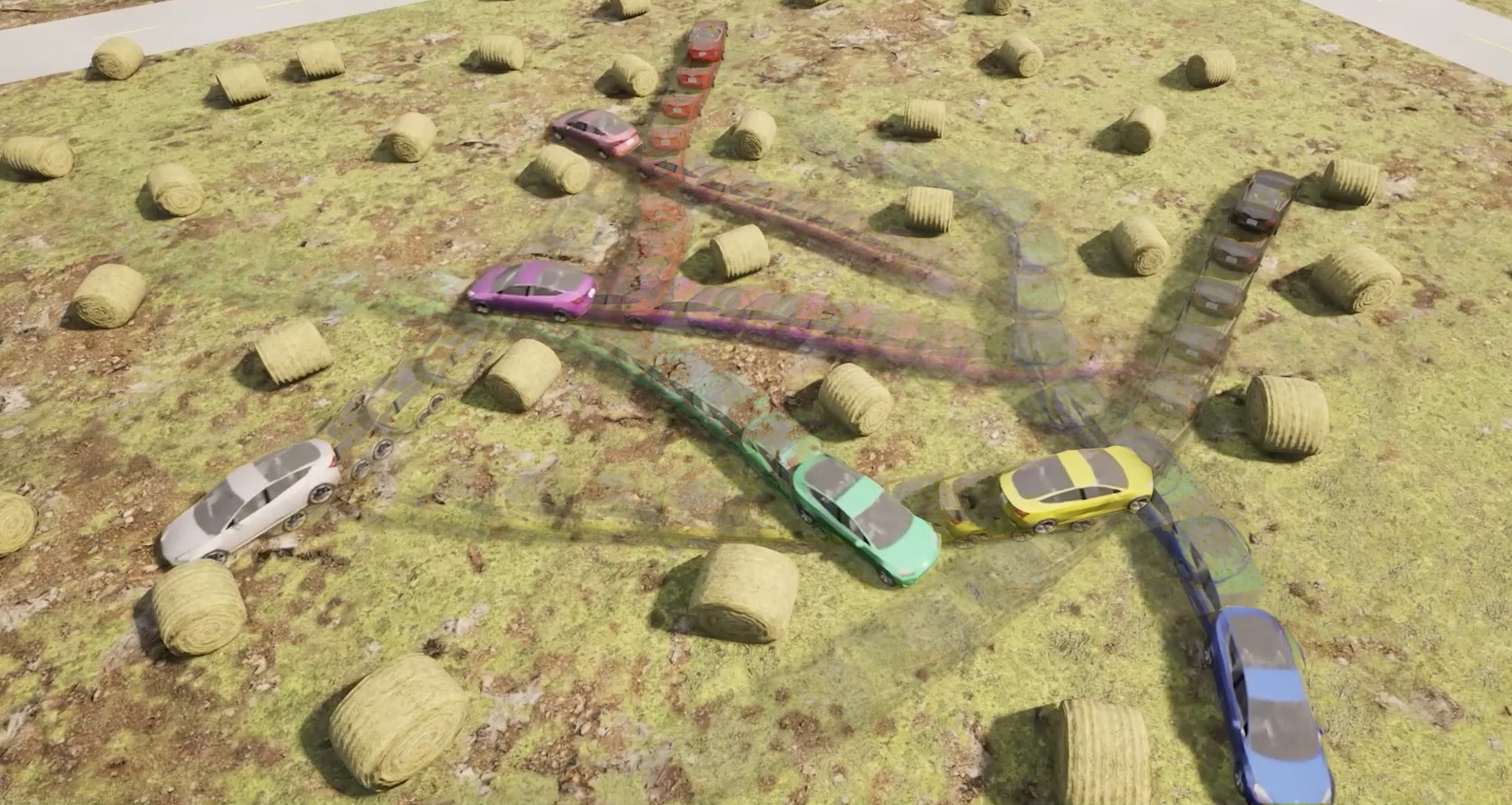}
    \caption{Car swarm in simulation environment}
    \label{fig:HeadFigure}
    \vspace{-0.65cm}
\end{figure}
In the back-end optimization, initial values play an important role.
Feasible initial values bring benefits such as speeding up optimization convergence and avoiding falling into a bad solution's local minima, whereas infeasible initial time arrangement may lead to convergence to a solution where the trajectory is in collision with other agents.
In order to acquire reasonable initial time allocation, we introduce search-based speed planning.
Once the spatial path is determined by the topology-guided path planning, the speed planning generates a time span that realizes collision avoidance with other agents as well as dynamic feasibility.
%Having determined the spatial path from topology-guided path planning, in order to acquire reasonable initial time allocation, we introduce search-based speed planning, which realizes collision avoidance with other agents as well as dynamic feasibility. 

Based on the spatial and temporal initial values, we employ a spatial-temporal joint optimization to generate a smooth, safe and feasible trajectory satisfying the non-holonomic constraints of car-like robots.
The proposed system is an extension and modification of our previous work \cite{han2022differential}, which focuses on single-car motion planning.
In addition to the topology-guided path planning and speed planning, we adopt a fixed time step trajectory discretization strategy to further improve the computation efficiency specifically for multi-agent missions.
%By discretizing the trajectory with fixed time steps, the computation efficiency is further improved.
In this system, trajectories are broadcast by a communication module, which will be utilized by other agents to perform real-time planning. 
The whole swarm system can traverse unknown environments while ensuring no collision between agents.

We validate the robustness of our system in a high-fidelity simulation platform for vehicles\cite{Dosovitskiy17} and real-world experiments. 
Each agent in this swarm independently carries out online sensing, mapping, planning and control, with trajectories broadcast to each other. Contributions of this paper are summarized as below:
\begin{enumerate}
    \item We introduce topology-guided path planning to improve the motion consistency of the robots, and leverage search-based speed planning to provide feasible initial values for the back-end optimization.
    \item We modify the trajectory discretization strategy with fixed time steps, improving the computation efficiency of the trajectory optimization process.
    \item We combine path planning, speed planning with spatial-temporal optimization, and implement the decentralized car-like robotic swarm, realizing real-time execution in cluttered environments.
    \item We will release our code as open-source packages to serve the research community\footnote{https://github.com/ZJU-FAST-Lab/Car-like-Robotic-swarm}.
\end{enumerate}

\section{RELATED WORKS}
\subsection{Trajectory Optimization for Car-Like Robots}
The complexity of kinematic models brings difficulties for planning on car-like robots, and optimization-based methods have been deeply explored to generate feasible trajectories. 
Zhu et al.\cite{zhu2015convex} transform the initial non-convex problem into a convex optimization with quadratic objective and quadratic constraints. 
Zhou et al.\cite{zhou2020autonomous} use sequential convex optimization to relax the curvature constraint.
Zhang et al.\cite{zhang2020optimization} propose an optimization-based collision avoidance algorithm.
Li et al.\cite{li2021optimization} realize collision avoidance by constraining the simplified car model in a safety corridor.
However, the above methods still suffer from the complicated mathematical model, resulting in low computation efficiency.
Recently, a differential flatness-based trajectory planning framework\cite{han2022differential} for vehicles is proposed which accomplishes fast computation in complicated environments. 
Based on the above baseline\cite{han2022differential}, we modified the trajectory discretization strategy to further decrease the computation burden in multi-robot scenarios.
\subsection{Car-like Robotic Swarm in Cluttered Environments}
Car-like robotic swarm can be divided as centralized and decentralized frameworks.
Mora et al.\cite{alonso2018cooperative,alonso2012reciprocal} realize collision avoidance based on RVO (Reciprocal Velocity Obstacles) and ORCA (Optimal Reciprocal Collision Avoidance), and then an optimization process is employed to generate smooth trajectories. 
Li et al.\cite{9812126,9345421} solve the time-optimal MVTP (Multi-Vehicle Trajectory Planning) problems in two stages and achieve swarm planning in dense environments. 
The above frameworks are based on centralized methods, while centralized methods are stuck by the computation complexity with the increase of the robot number. 
Delimpaltadaki et al.\cite{8258896} propose a decentralized framework to solve the predecessor-following control problem. 
However, this work\cite{8258896} is limited to platooning scenarios without adaptability to complicated tasks such as crossovers.
In this paper, we adopt the decentralized planning framework to construct car-like robotic swarm.

\subsection{Topology-Guided Path Planning}
Topological planning has been applied in many scenarios to choose the best spatial path and avoid local minima. Jaillet et al.\cite{jaillet2008path} construct visibility deformation roadmaps to define the topological homotopy and encode the environmental topological information. 
Zhou et al.\cite{zhou2020robust, zhou2021raptor} extend visibility deformation to propose real-time planning by checking topological equivalence.
Zhou et al.\cite{zhou2021ego} construct distance fields in different directions of obstacles and utilize visibility deformation to find distinctive trajectories. 
In this paper, we extend visibility deformation in works\cite{jaillet2008path,zhou2020robust} to a broader definition to search for a path homeomorphic with the last planned path.

\subsection{Speed Planning}
Spatial-temporal decomposition is a widely used strategy in autonomous driving to improve planning efficiency, where speed planning is the temporal part. 
Works\cite{cheng2022real,fan2018baidu} use search-based speed planning to search for a feasible speed profile. 
Works\cite{liu2017speed, xu2022speed} propose optimization-based speed planning methods to generate a smooth S-T (Space-Time) curve. 
Johnson et al.\cite{johnson2012optimal,johnson2013optimal} realize dynamic obstacle avoidance by propagating reachable velocity sets. 
Other works\cite{li2021speed,gonzalez2016speed} conduct speed planning by optimizing Bézier Polynomials. 
However, most existing methods focus on getting a smooth and feasible speed profile without spatial-temporal joint optimization.
In this paper, we decouple space-time planning, where initial time span is provided by the search-based speed planning. 
Then the initial path is refined by subsequent spatial-temporal optimization.
Therefore, the initial time span is not necessary to be as smooth or accurate as existing methods require.

\section{Path and speed planning} 
In this section, we will introduce topology-guided path planning and search-based speed planning, which serve as spatial and temporal initial values for the back-end optimization.\looseness=-1

\begin{algorithm}[htpb]
	\caption{$SD$-VD guided path expansion}
	% \scriptsize
	\small
	\label{alg:path expansion}
	\textbf{Notation}: Previously planned path $\tau_0$; Newly planned path $\tau$; Arc length of each motion primitive $\Delta{s}$; Set of newly expanded primitives from one node $\mathcal{Q}$;  \\
	\KwIn{
		$\tau_0$, $\tau(0)$
	}
	\KwOut{
		$\tau$
	} 
	\BlankLine
	Initialize: $\tau_0(s_0)=\mathrm{FindNearestNode}(\tau(0),\tau_0)$, $k=0$ \label{alg:pathexpansion_init}\;
	\While{$k\Delta{s} \leq S$}{\label{alg:expand_while}
		$\mathcal{Q}=\mathrm{expandNodes}(\tau(k\Delta{s}))$\;
		\For{$q_k:\mathcal{Q}$}{\label{alg:for loop}
			$p^{k+1}=\mathrm{getPos}\left(q_k,\tau((k+1)\Delta{s} ) \right)$\;
			$p_0^{k+1}=\mathrm{getPos}\left(q_k,\tau_0(s_0+(k+1)\Delta{s}) \right)$\;
			NotCol = $\mathrm{checkCollisionUsingPoint}(p^{k+1})$\;
			NotOcc = $\mathrm{checkCollisionUsingLine}(p^{k+1},p_0^{k+1})$\;
			IsWithinDis = $\mathrm{checkDistance}(p^{k+1},p_0^{k+1},D)$\;
			\If(){$\mathrm{NotCol}$ and $\mathrm{NotOcc}$ and $\mathrm{IsWithinDis}$}{
				$\tau$.pushback($q_k$);
			}
		}
		$k=k+1$\;
	}\label{alg:samp_while_end}% END of WHILE
\end{algorithm}

\subsection{Topology-Guided Path Planning}
To generate a spatial reference path before optimization, we adopt the Kinodynamic hybrid A*\cite{dolgov2010path} on the car case. Instead of sampling different poses, we directly sample different steering angles, and each motion primitive shares the same longitude distance. By limiting the maximum steering angle, the non-holonomic constraints can be satisfied initially.\looseness=-1

The concept of topological homotopy has been analysed in \cite{jaillet2008path,zhou2020robust}, which aims to capture candidate trajectories with different topological structures. 
However, both visibility deformation (VD) in \cite{jaillet2008path} and uniform visibility deformation (UVD) in \cite{zhou2020robust} require that candidate trajectories should share the same start and end point. 
In our searching process, replan start point may not be located on the previously searched path, hence we extend VD to a broader definition as $SD$-VD below:\looseness=-1

\textbf{\textit{Definition 1:}} $SD$-VD condition: for pre-defined distances $S, D\in \mathbb{R}^+$, two trajectories $\tau_1(s), \tau_2(s): \mathbb{R} \mapsto \mathbb{R}^2$, parameterized by arc length $s\geq0$, belong to the same $SD$-VD class if for all $s\leq S$, line $\tau_1(s)-\tau_2(s)$ is collision-free, and $||\tau_1(s)-\tau_2(s)||_2\leq D$.

We introduce $SD$-VD into our path expansion process to get a kinodynamic path homeomorphic with the last planned one, as shown in Alg.\ref{alg:path expansion}. 
Before path expansion, the previous kinodynamic path is stored as $\tau_0(s): \mathbb{R} \mapsto \mathbb{R}^2$, representing the position on the path $\tau_0$ at the arc length of $s$.
The newly planned path is noted as $\tau(s): \mathbb{R} \mapsto \mathbb{R}^2$, so the replan start point is $\tau(0) \in \mathbb{R}^2$.
Each expanded primitive shares the same arc length $\Delta{s}$.
At initialization in Line 2, iterate over all points in $\tau_0$ and find the point $\tau_0(s_0) \in \mathbb{R}^2$ which is the nearest one to the replan start point $\tau(0)$.
During the expansion, in Line 4, all motion primitives expanded from the $k$-th node $\tau(k\Delta{s})$ are stored in vector $\mathcal{Q}$. 
In Lines 5-14, the primitives in vector $\mathcal{Q}$ are iterated to check if they satisfy the $SD$-VD conditions. 
In Line 6-7, the function \text{getPos()} returns the position at the arc length of $(k+1)\Delta{s}$ and $s_0+(k+1)\Delta{s}$ on $\tau$ and $\tau_0$, respectively noted as $p^{k+1}\in \mathbb{R}^2$ and $p_0^{k+1}\in\mathbb{R}^2$. 
In Lines 8-10, the main conditions of $SD$-VD are checked. 
In Line 8, \text{NotCol} is true if $p^{k+1}$ is not colliding with the obstacles; In Line 9, \text{NotOcc} is true if line $p^{k+1}-p_0^{k+1}$ is not intersecting with any obstacle; In Line 10, \text{IsWithinDis} is true if the distance between $p^{k+1}$ and $p_0^{k+1}$ is smaller than $D$.
Only if all the above conditions are satisfied will the $SD$-VD condition be satisfied and the expanded node be considered as a valid path to push into $\tau$.

\begin{figure}[t] 
    \centering
    \includegraphics[width=8.8cm]{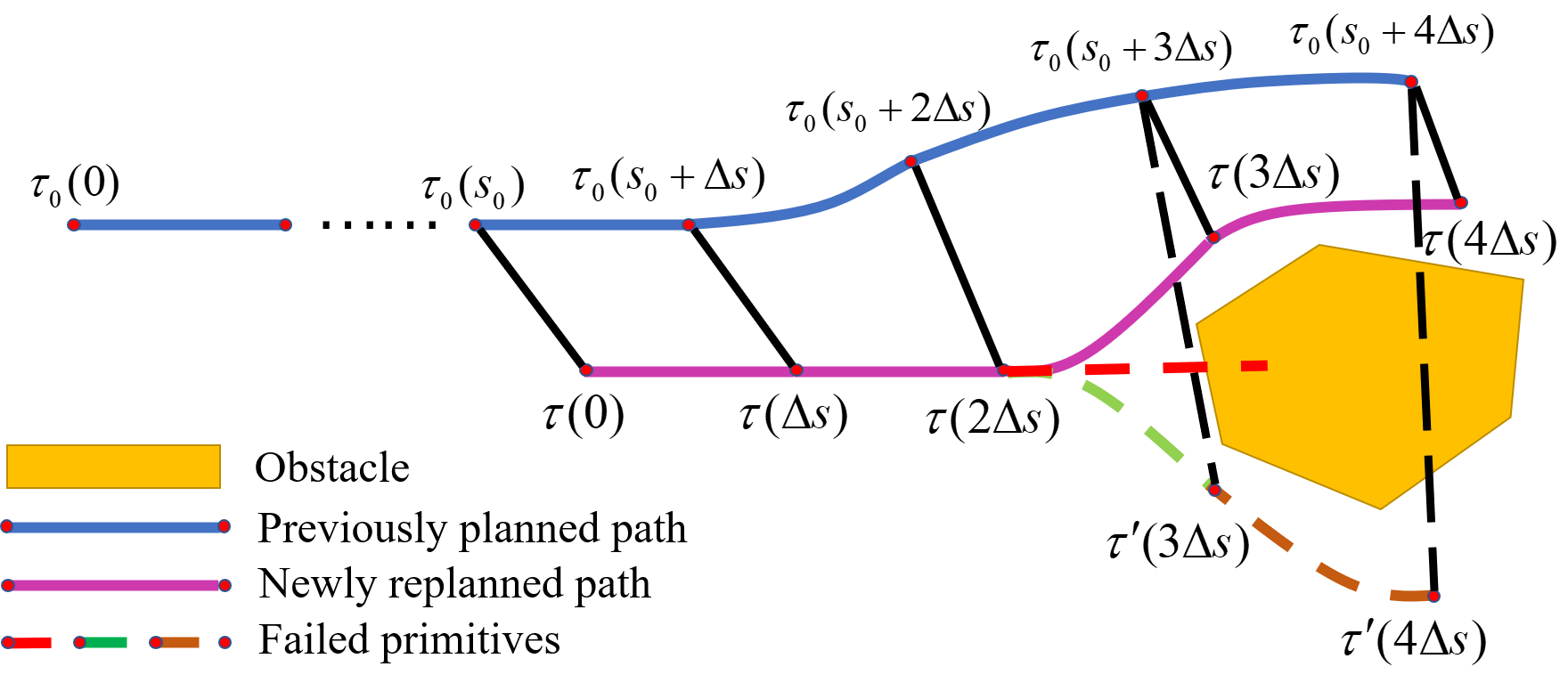}
    \setlength{\abovecaptionskip}{-0.5cm}
    \caption{This figure illustrates the topology-guided path planning. The purple curve is the newly replanned path in homotopy with the previously planned path. Whereas the green and brown dotted line is not in the $SD$-VD class with the blue line, therefore abandoned.}
    \label{fig: Topology-guided path planning}
    \vspace{-0.5cm}
\end{figure}

The $SD$-VD and searching process are shown in Fig.\ref{fig: Topology-guided path planning}.
Distances $S$ and $D$ are chosen based on the sensing range and the robot's kinematic properties, which demand that there is enough space for the robot to adjust the steering angle without colliding with obstacles. 
Subsequently, the searched path lies in the same topological homotopy as the last planned path, guaranteeing trajectory consistency and avoiding collisions.

% We introduce $SD$-VD into our path planning process to get a kinodynamic path homeomorphic with the last planned one. 
% Before path searching, the previous kinodynamic path is stored as $\tau_0(s)$. 
% When replan is triggered, each node point in $\tau_0(s)$ is iterated and the point $\tau_0(s_0)$ which is nearest to the replan start point is found out. 
% The replan start point is noted as $\tau(0)$. 
% Suppose the arc length of each motion primitive is $\Delta s$, and the position of the $k$-th primitive starting from the replan point is $\tau(k\Delta s)$. 
% After that, find the corresponding primitive of the previous path $\tau_0(s_0+k\Delta s)$, check if line $\tau(k\Delta s)$-$\tau_0(s_0+k\Delta s)$ is collision-free and if $||\tau(k\Delta s)-\tau_0(s_0+k\Delta s)||_2 \leq D$. 
% If either of the above requirements is not satisfied, the newly searched primitive and the previous path is not in the same $SD$-VD class, and this primitive will not be accessed anymore. 
% The SD-VD and searching process are shown in Fig.\ref{fig: Topology-guided path planning}.
% Distances $S$ and $D$ are chosen based on the sensing range and the robot's kinematic properties, which demands that there is enough space for the robot to adjust the steering angle without colliding with obstacles. 
% Subsequently, searched path lies in the same topological homotopy with the last planned path, guaranteeing trajectory consistency and avoiding collisions.

\subsection{Search-based Speed Planning}
Based on the spatial path generated in \uppercase\expandafter{\romannumeral3}-$A$ and trajectories broadcast from other agents, an S-T graph is constructed before speed planning. 
As done in \cite{fan2018baidu}, we discrete arc length space S and time space T into grids, and iterate points in S-T space to check if the ego robot is in collision with other agents. 
If colliding, the corresponding grid in the S-T graph is noted as a collision region, which means that this region is not accessible in speed planning.

Speed planning generates a time span satisfying dynamic obstacle avoidance and kinematic feasibility. 
We adopt a search-based speed planning method, similar to \cite{cheng2022real}, and conduct a 1-dimension A* search on the S-T graph. 
The control input is set as the tangential acceleration $\mathcal{A}=\left\{a_{\min } \leq a_{1}, \ldots, a_{k} \leq a_{\max }\right\}$, and each child node shares the same time interval $T_f$. 
Starting from the initial state $x_0$, the forward expansion is: 
\begin{equation}
\left[\begin{array}{c}
x_{i+1}^{j} \\
\dot{x}_{i+1}^{j}
\end{array}\right]=\left[\begin{array}{c}
x_{i}+\dot{x}_{i} T_{f}+\frac{1}{2} a_{j} T_{f}^{2} \\
\dot{x}_{i}+a_{j} T_{f}
\end{array}\right], 1 \leq j \leq k .
\end{equation}

\begin{figure}[htp]
	\centering
	\includegraphics[width=8.8cm]{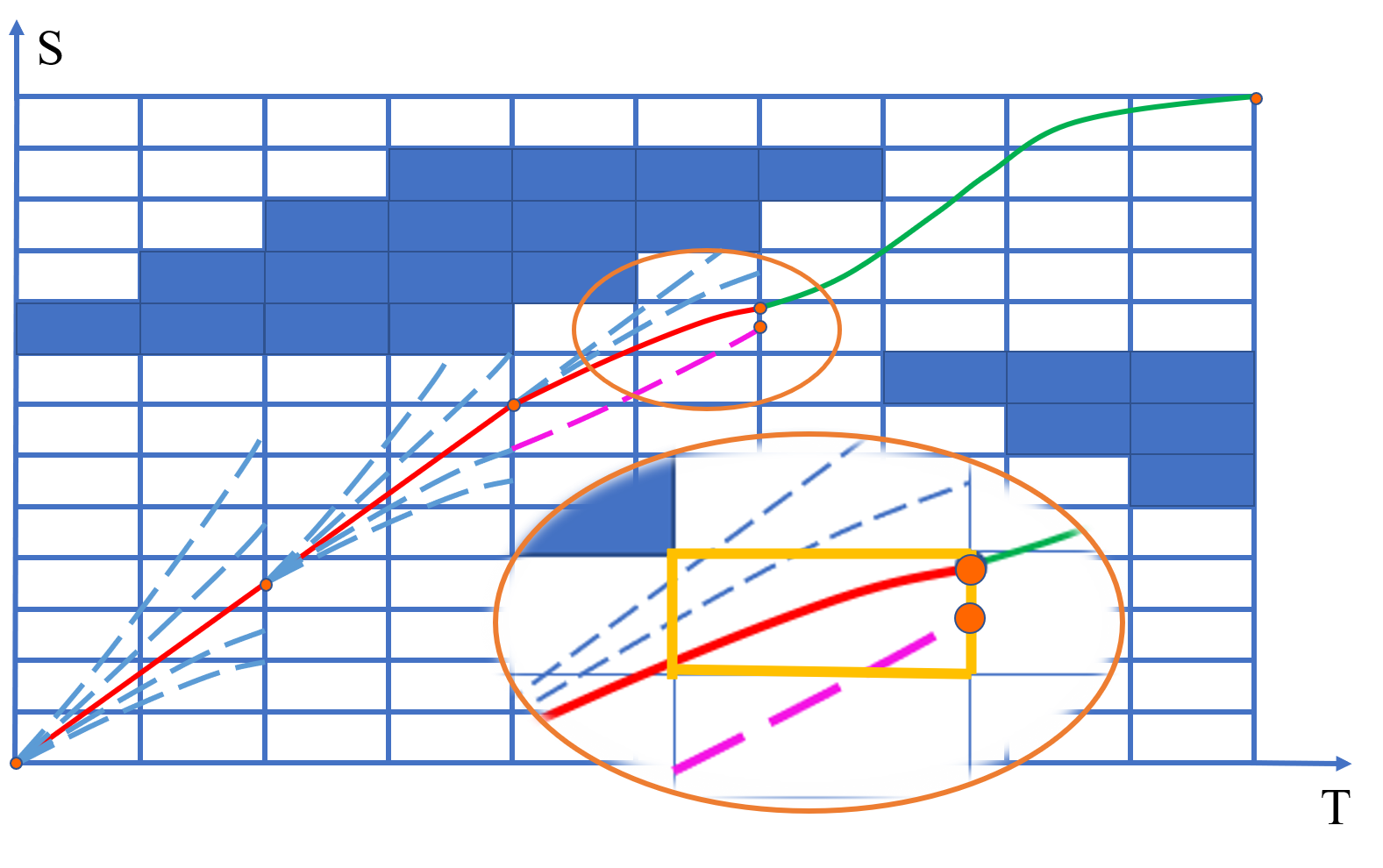}
	\setlength{\abovecaptionskip}{-0.6cm}
	\caption{Illustration of search-based speed planning on S-T graph. Red curves are the adopted nodes, and the green curve is the one-shot curve. The blue dotted lines are abandoned because of collision with obstacles or dynamical infeasibility. The red curve and the purple curve lie in the same S-T-V grid, and the purple node's cost is higher, hence this node is pruned.}
	\label{fig: speed planning}
	\vspace{-0.45cm}
\end{figure}
During the searching process, any node that violates the dynamic constraints or reaches the collision regions will be abandoned. 
The cost of each node is defined as:
\begin{equation}
J_{p}=J_{g}+\lambda_h J_{h},
\end{equation}
where $J_g$ means the time spent from the start state $x_0$ to the current state, $J_h$ represents the minimum time from the current state to the end state, and $\lambda_h$ is the weight parameter. 

Note that the number of node states is growing rapidly during expansion, which suppresses the planner's real-time performance. To solve this problem, we adopt two strategies. 

1) \textit{Node pruning mechanism:} In S-T graph construction, we discrete S space and T space. Based on that, we add velocity space discretization, thus each node state belongs to an S-T-V discretization grid. If the newly expanded node falls into the same S-T-V grid as one expanded node before, we only reserve the node with lower cost $J_p$, and the other is pruned, as shown in Fig.\ref{fig: speed planning}. By restricting S-T-V space discretization, the number of redundant nodes can be remarkably decreased. 

2) \textit{One-shot mechanism:} While accessing each expanded node, perform bang-bang control\cite{doi:10.1137/1022026} directly from the current state to the end state. If the bang-bang control curve on the S-T graph doesn't intersect with the collision regions, the searching process is terminated and corresponding nodes serve as the final result of the speed planning. By applying the one-shot mechanism, the searching process can finish in advance, which guarantees real-time performance.

\section{Spatial-Temporal optimization}
In this section, we introduce the trajectory representation and the optimization formulation. 
To calculate the cost function, we adopt the fixed time discretization strategy, from which the computation burden is further ameliorated. 
To satisfy the non-holonomic constraints, the mathematical deviations of differential flatness cooperated with the constraints on the front steer angle are introduced as well.
Then we introduce the formulation of the collision avoidance between agents in swarm scenarios.
After that, the advantages of the adopted trajectory discretization strategy are discussed.
% For the integrity of this framework, we introduce the main mathematical deviations of optimization formulation, steer angle limitation and dynamic agent avoidance.

\subsection{Trajectory and Optimization Formulation}
We utilize differential flatness property for the simplified kinematic bicycle model in the Cartesian coordinate to describe the four-wheel vehicle. As depicted in \cite{fuchshumer2005nonlinear}, states of car-like robots can be derived from the flat output $\boldsymbol{p}:=\left[p_{x}, p_{y}\right]^{T}$, where $\boldsymbol{p}$ means the position centered on rear wheel of the car. 

Trajectory is formulated by the 2-dimensional polynomial with degree $N=2m-1$. 
In our framework we select $m=3$. Suppose a segment of the trajectory consists of $M$ pieces, and pieces in a segment are time-uniform, where the time interval for each piece is $\delta T \in \mathbb{R}^+$, so the overall duration of the segment is $\mathrm{T}=M\times \delta T$. 
For the $i$-th piece, the coefficient vector is $\mathbf{c}_i \in \mathbb{R}^{2m\times 1}$, then this piece can be written as:   
\begin{align}
	\label{equ: trajectory representation}
\begin{gathered}
\boldsymbol{p}_{i}(t) :=\mathbf{c}_{i}^T \boldsymbol{\beta}(t), \\
\boldsymbol{\beta}(t) :=\left[1, t, t^2, \ldots, t^N\right]^T,
\end{gathered}
\end{align}  
where $t\in [0, \delta T]$, and $i\in \left\{1,2,...,M \right\}$.

Before we introduce the optimization formulation, we first present the violation function $\mathcal{G}_d$, where $d \in \mathcal{D}$, and $\mathcal{D}$ is the set of constraint terms such as dynamic feasibility, front steer angle limitation, static and dynamic obstacles collision avoidance. The mathematical description of the constraint terms is essentially inequality constraints:
\begin{align}
	\label{equ: inequality constraints}
\mathcal{G}_d\left(\boldsymbol{p}_{i}(\bar{t}), \ldots, \boldsymbol{p}_{i}^{(m)}(\bar{t})\right) \preceq \mathbf{0},
\end{align}
where $\bar{t}$ is the relative timestamp of some constraint point in this piece.

It is proved in \cite{wang2022geometrically} that the inequality constraints in Eq.(\ref{equ: inequality constraints}) can be transformed into a penalty term $S_{\Sigma}$, and the formulation is an unconstrained nonlinear optimization problem:
\begin{align}
\min _{\mathbf{c}, \mathrm{T}} &J=\int_0^{\mathrm{T}} \boldsymbol{\mu}(t)^T \boldsymbol{\mu}(t) d t+w_T \mathrm{T} +S_{\Sigma}(\mathbf{c}, \mathrm{T}),
\end{align}
where $\mathbf{c} = \left[ \mathbf{c}_1,...,\mathbf{c}_M \right] \in \mathbb{R}^{2m \times M}$ is the coefficient matrix. $\boldsymbol{\mu}(t)$ denotes the control efforts $\boldsymbol{p}^{(m)}$, and $w_T$ represents the penalty weight on total time $\mathrm{T}$.

The trajectory is discretized into multiple constraint points by fixed time steps, instead of the fixed number in a piece. Each time step is $\delta t=0.2s$, and the total number of constraint points is $K=\left\lfloor\mathrm{T} /\delta t\right\rfloor$. In order to satisfy the inequality constraints, we penalize all constraint points and integrate the violation function over time to get the penalty term. The expression of the penalty is: 
\begin{align}
	&S_{\Sigma}=\sum_{d \in \mathcal{D}} w_d \sum_{i=1}^M \sum_{j=1}^{K} P_{d, i, j}\left(\mathbf{c}_{i}, \mathrm{T}\right), \\
	&P_{d, i, j}\left(\mathbf{c}_{i}, \mathrm{T}\right)=\delta t  \mathrm{~L}_1\left(\mathcal{G}_{d, i, j}\right), 
\end{align}

where $w_d$ is the penalty weight, and $\mathrm{L}_1(\cdot)$ is the L1-norm relaxation function. 
Suppose the $j$-th constraint point is located at the $i$-th piece, then we have:
\begin{align}
	\bar{t}=j \delta t - i \frac{\mathrm{T}}{M}.
\end{align}
According to the chain rule, the gradients of $J$ w.r.t $c_{i}$ and $T$ are converted to the gradients of $\mathcal{G}_d$ w.r.t $\boldsymbol{p}$,...,$\boldsymbol{p}^{(m)}$and $\bar{t}$:
\begin{gather}
	\frac{\partial P_{d,i,j}}{\partial c_i}=\delta t \mathrm{~L}_{1}^{\prime}(\mathcal{G}_{d,i,j})
	\left[ \sum_{d=0}^m \beta^{(d)}(\bar{t})\left(\frac{\partial \mathcal{G}_{d,i,j}}{\partial \boldsymbol{p}^{(d)}}\right) \right], \\
	\frac{\partial P_{d,i,j}}{\partial \mathrm{T}}=\delta t \mathrm{~L}_{1}^{\prime}(\mathcal{G}_{d,i,j})
	\left[\frac{\partial \mathcal{G}_{d,i,j}}{\partial \bar{t}} \cdot \frac{\partial \bar{t}}{\partial \mathrm{T}} \right], \\
	\frac{\partial \bar{t}}{\partial \mathrm{T}}=-\frac{i}{M}.
\end{gather}

Next we will introduce the mathematical formulation of the front steer constraints and dynamic obstacle avoidance.

\subsection{Front Steer Angle Limitation}
The non-holonomic constraints of car-like robots is associated with the front steer angle limitation. According to the differential flatness model, the steering angle $\phi$ can be expressed by the flat output:
\begin{equation}
\phi=\arctan \left(\eta\left(\dot{p}_x \ddot{p}_y-\dot{p}_y \ddot{p}_x\right) L /\left({\dot{p_x}}^2+{\dot{p_y}}^2\right)^{\frac{3}{2}}\right),
\end{equation}
where $\eta \in \left\{-1, 1 \right\}$, representing the car is moving forward or backward. $L$ is the wheelbase length. Then the curvature $\mathcal{C}$ is: 
\begin{equation}
\mathcal{C} = \tan{\phi} / L = \frac{\ddot{\boldsymbol{p}}^T \mathbf{H} \dot{\boldsymbol{p}}}{\| \dot{\boldsymbol{p}}\|_2^3},
\end{equation}
where $\mathbf{H}:=\left[\begin{array}{cc}
0 & -1 \\
1 & 0
\end{array}\right]$. Suppose the maximum curvature is $\mathcal{C}_m$, then the violation function is:
\begin{equation}
\mathcal{G}_{\mathcal{C}}(\dot{\boldsymbol{p}}, \ddot{\boldsymbol{p}}) = 
\frac{\ddot{\boldsymbol{p}}^T \mathbf{H} \dot{\boldsymbol{p}}}{\| \dot{\boldsymbol{p}}\|_2^3} - \mathcal{C}_m.
\end{equation}

Then the gradients can be naturally derived as: 
\begin{align}
\frac{\partial \mathcal{G}_{\mathcal{C}}}{\partial \dot{\boldsymbol{p}}}&=\frac{\mathbf{H}^T \ddot{\boldsymbol{p}}}{\|\dot{\boldsymbol{p}}\|_2^3}-3 \frac{\ddot{\boldsymbol{p}}^T \mathbf{H} \dot{\boldsymbol{p}}}{\|\dot{\boldsymbol{p}}\|_2^5} \dot{\boldsymbol{p}},\\
\frac{\partial \mathcal{G}_{\mathcal{C}}}{\partial \ddot{\boldsymbol{p}}}&=\frac{\mathbf{H} \dot{\boldsymbol{p}}}{\|\dot{\boldsymbol{p}}\|_2^3},\\
\frac{\partial{\mathcal{G}_{\mathcal{C}}}}{\partial{\bar{t}}} &= \frac{\partial \mathcal{G}_{\mathcal{C}}}{\partial \dot{\boldsymbol{p}}} \frac{\partial{\dot{\boldsymbol{p}}}}{\partial{\bar{t}}} + 
\frac{\partial \mathcal{G}_{\mathcal{C}}}{\partial \ddot{\boldsymbol{p}}} \frac{\partial{\ddot{\boldsymbol{p}}}}{\partial{\bar{t}}}.
\end{align}

\subsection{Collision Avoidance between Agents}
In this system, each agent is modeled as a convex polygon. In the optimization process, dynamical obstacle collision avoidance is satisfied by calculating the signed distances at each constraint point between convex polygons, and penalty is imposed if the distances are short.

Suppose $n_e$ and $n_o$ respectively represent the number of the vertexes of ego car and obstacle polygon, and the coordinates of the $e$-th and $o$-th vertex of ego car and obstacle car are $\boldsymbol{v}_{E}^{e}$ and $\boldsymbol{v}_{O}^{o}$, where $1 \leq e \leq n_e$, $1 \leq o \leq n_o$. The index of the vertexes is counted clockwise. So the distance from the $o$-th vertex of the obstacle to the $e$-th edge of ego car is: 
\begin{align}
	\label{equ: d_e_o}
d_{e}^{o} = \frac{\left[ \mathbf{H}(\boldsymbol{v}_E^{e+1} - \boldsymbol{v}_{E}^{e}) \right]^T (\boldsymbol{v}_{O}^{o} - \boldsymbol{v}_{E}^{e})}{\| \boldsymbol{v}_{E}^{e+1} - \boldsymbol{v}_{E}^{e} \|_2}.
\end{align}
\begin{figure}[htp]
	\centering
	\includegraphics[width=8.7cm]{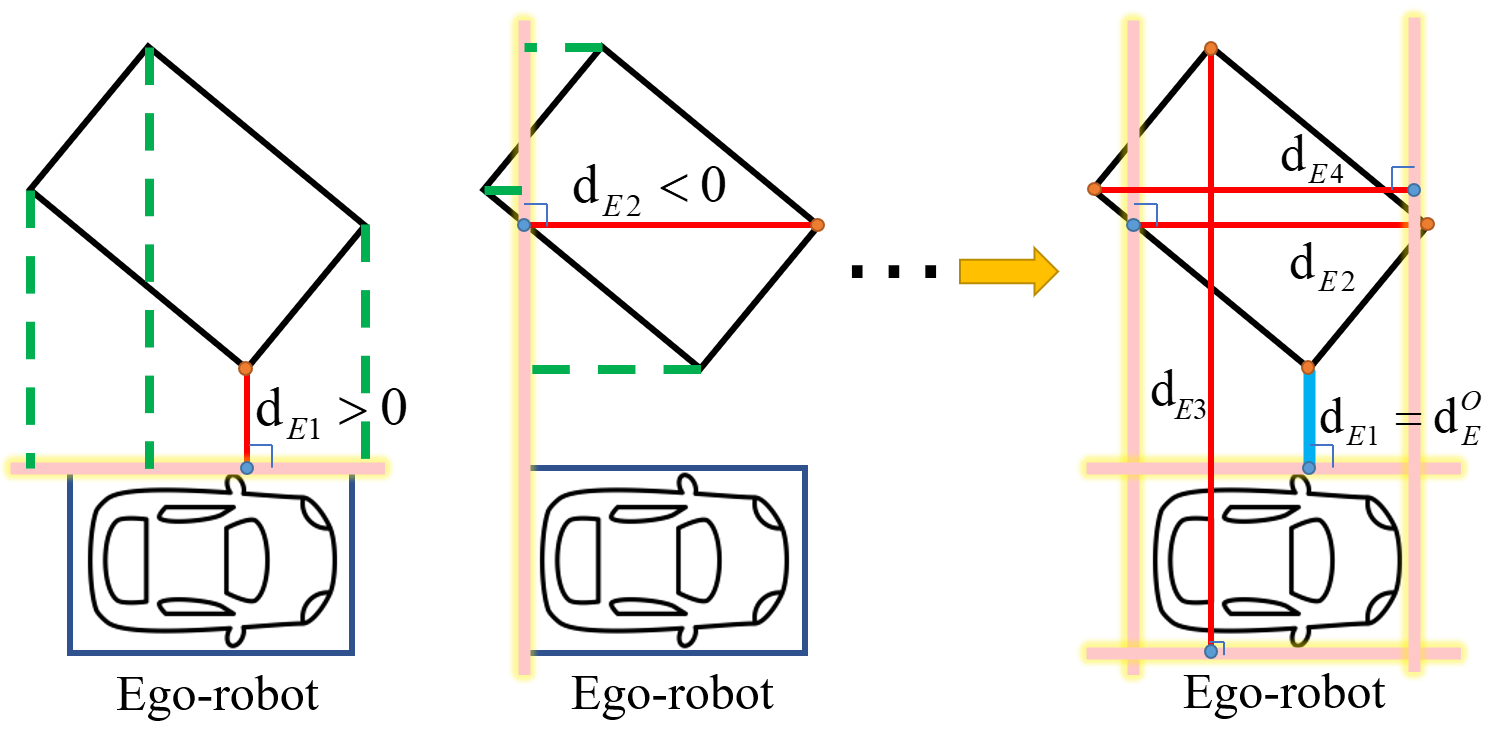}
	\setlength{\abovecaptionskip}{-0.6cm}
	\caption{Illustration of signed distance from other agent to ego robot.}
	\label{fig: signed distance}
	\vspace{-0.5cm}
\end{figure}
For the $e$-th edge of ego car, iterate over all vertexes of obstacle and find the minimum distance:
\begin{align}
d_{Ee} \approx \mathrm{lse}_{\alpha \textless 0} \left( \left[ d_{e}^{1},...,d_{e}^{o},...,d_{e}^{n_o} \right]^T \right).
\end{align}
Then we choose the maximum value of $d_e$ as the signed distance from the obstacle to ego car:
\begin{align}
\mathbf{d}_{E}^{O} \approx \mathrm{lse}_{\alpha \textgreater 0} \left( \left[ d_{E1},...,d_{Ee},...,d_{E n_e}  \right]^T \right).
\end{align}
Likewise, the signed distance from the ego car to the obstacle is $\mathbf{d}_{O}^{E}$.
Note that a log-sum-exp function is leveraged to approximate and smooth the $\max$ and $\min$ functions. If $\alpha \textgreater 0$, the function returns the maximum value in the input vector, and vice versa.
Finally, the signed distance between ego car and obstacle is: 
\begin{align}
\mathbf{d} \approx \mathrm{lse}_{\alpha \textless 0} \left( \left[ \mathbf{d}_{E}^{O}, \mathbf{d}_{O}^{E} \right]^T \right).
\end{align} 
The signed distance is shown in Fig.\ref{fig: signed distance}.

Based on the signed distance above, the violation function of collision $\mathcal{G}_{EO}$ is defined as: 
\begin{align}
\mathcal{G}_{EO}(\boldsymbol{p}, \dot{\boldsymbol{p}}, \bar{t}) = d_{tol} - \mathbf{d},
\end{align}
where $d_{tol}$ is the safety margin distance. According to the chain rule:
\begin{align}
\frac{\partial \mathcal{G}_{EO}}{\partial \boldsymbol{p}} = -\mathrm{lse}_{\alpha \textless 0}^{\prime}(\mathbf{d}_{E}^{O}) \frac{\partial \mathbf{d}_{E}^{O}}{\partial \boldsymbol{p}} - \mathrm{lse}_{\alpha \textless 0}^{\prime}(\mathbf{d}_{O}^{E}) \frac{\partial \mathbf{d}_{O}^{E}}{\partial \boldsymbol{p}}, \\
\frac{\partial \mathbf{d}_{E}^{O}}{\partial \boldsymbol{p}} = \sum_{e=1}^{n_e}\mathrm{lse}_{\alpha \textgreater 0}^{\prime}(d_{Ee})\left( \sum_{o=1}^{n_o}\mathrm{lse}_{\alpha \textless 0}^{\prime}(d_e^o)\frac{\partial d_e^o}{\partial \boldsymbol{p}} \right), 
\end{align}
Similarly, the gradient of $\mathcal{G}_{EO}$ w.r.t $\dot{\boldsymbol{p}}$ and $\bar{t}$ are mainly about the gradient of $d_e^o$ and $d_o^e$ w.r.t $\dot{\boldsymbol{p}}$ and $\bar{t}$. 
Suppose the $e$-th vertex of ego car's relative coordinate to body frame is $\boldsymbol{l}_E^e$, the rotation matrix is $\mathbf{R}$, and the vertex $\boldsymbol{v}_E^e = \boldsymbol{p} + \mathbf{R}\boldsymbol{l}_E^e$. Substitute $\boldsymbol{v}_{E}^e, \boldsymbol{v}_{O}^o$ in Eq.(\ref{equ: d_e_o}) we can get:
\begin{align}
	\label{equ: signed distance}
d_e^o = \frac{\left[ \mathbf{HR}(\boldsymbol{l}_E^{e+1} - \boldsymbol{l}_E^e) \right]^T}{\|\boldsymbol{l}_E^{e+1} - \boldsymbol{l}_E^e \|_2}\left( \hat{\mathbf{R}}\boldsymbol{l}_O^o + \hat{\boldsymbol{p}} - \mathbf{R}\boldsymbol{l}_E^e - \boldsymbol{p} \right),
\end{align}
where $\hat{\mathbf{R}}$ and $\hat{\boldsymbol{p}}$ represent the rotation matrix and position of the obstacle, respectively. In this equation, $\mathbf{R}$ is associated with $\boldsymbol{p}$, $\dot{\boldsymbol{p}}$ and $\bar{t}$. Therefore, the gradients of $d_e^o$ w.r.t $\boldsymbol{p}, \dot{\boldsymbol{p}}, \bar{t}$ are further converted to the gradients of $\mathbf{R}(\boldsymbol{p}, \dot{\boldsymbol{p}}, \bar{t})$ w.r.t $\boldsymbol{p}, \dot{\boldsymbol{p}}, \bar{t}$. 

\subsection{Discussion}
\begin{figure}[htp]
	\centering
	\includegraphics[width=8.7cm]{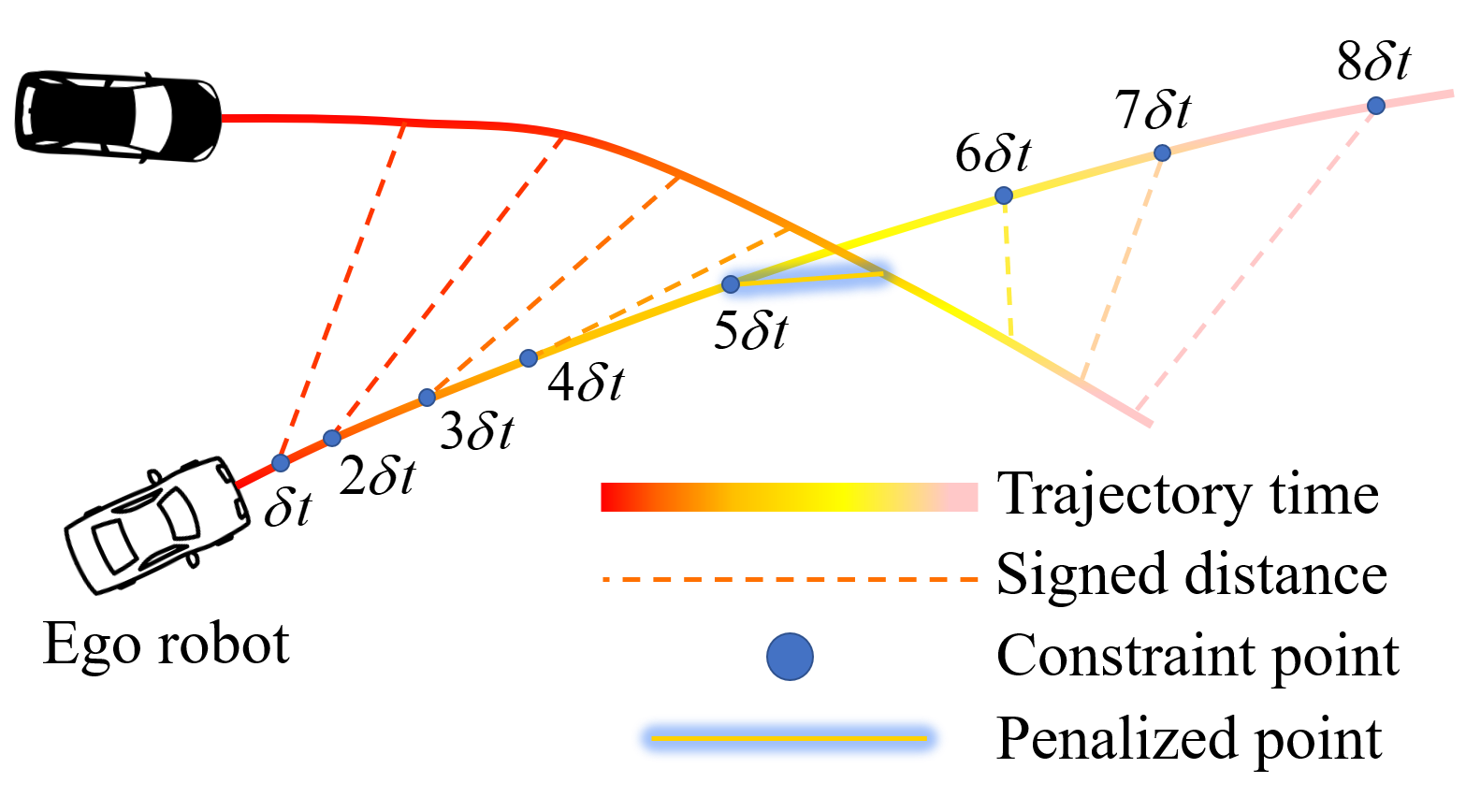}
	\setlength{\abovecaptionskip}{-0.6cm}
	\caption{Illustration of obstacle avoidance between agents. In each iteration of the optimization process, the trajectory is discretized by a fixed time step $\delta t$. At each constraint point, the signed distance between ego robot and other agent is calculated. If the signed distance is less than $d_{tol}$, for example, at the timestamp $5\delta t$ in this figure, the penalty will be imposed and the gradients will be propagated to further refine the trajectory.}
	\label{fig: dynamic obstacle avoidance}
	\vspace{-0.5cm}
\end{figure}
Here we discuss about the advantage of the fixed time step discretization strategy. 
In our previous work\cite{han2022differential}, each piece of the trajectory is uniformly discretized by the fixed number of constraint points. 
With the optimization process iterating, the positions and timestamps of the constraint points are changing. 
To acquire the states of dynamic obstacles at the moment of each constraint point, we introduce $\hat{t}$ which means the absolute timestamp of the constraint point. 
Intuitively, $\bar{t}$ is used to calculate the states of the ego robot, whereas $\hat{t}$ is to index and calculate the states of other agents via the communicated trajectories. 
Both $\bar{t}$ and $\hat{t}$ are associated with the total time duration of the trajectory $\mathrm{T}$.

In Eq.(\ref{equ: signed distance}), variables $\hat{\mathbf{R}}$ and $\hat{\boldsymbol{p}}$ are related to the absolute timestamp $\hat{t}$. 
In our previous work, the violation function is defined as $\mathcal{G}_{EO}(\boldsymbol{p},\boldsymbol{\dot{p}},\bar{t},\hat{t})$. 
In order to get $\partial \mathcal{G}_{EO} / \partial \hat{t}$, the gradients $\partial \hat{\mathbf{R}}/\partial \hat{t}$ and $\partial \hat{\boldsymbol{p}} /\partial \hat{t}$ should be considered and calculated as well.  
In this paper, we discretize the trajectory with fixed time steps, which means that the timestamps of the constraint points are not varying with the optimization iterations.
Therefore, the states of other agents $\hat{\mathbf{R}}$ and $\hat{\boldsymbol{p}}$ corresponding to each constraint point are fixed, indicating that the extra variable $\hat{t}$ and related calculations can be omitted.
In addition, before a trajectory starts to be optimized, the states $\hat{\mathbf{R}}$ and $\hat{\boldsymbol{p}}$ can be calculated and stored into a list in advance.
Hence there is no need to calculate the states of other agents repeatedly during the optimization iterations.

The illustration of obstacle avoidance between agents is shown in Fig.\ref{fig: dynamic obstacle avoidance}.
Detailed comparisons will be present in Sec.\ref{sec: experiments}.
Due to the page limitations, we will not present more mathematical derivations such as dynamic feasibility and static obstacle avoidance. For more details, we refer readers to our previous work.

\section{System Architecture}
\begin{figure}[htp]
	\centering
	\includegraphics[width=8.7cm]{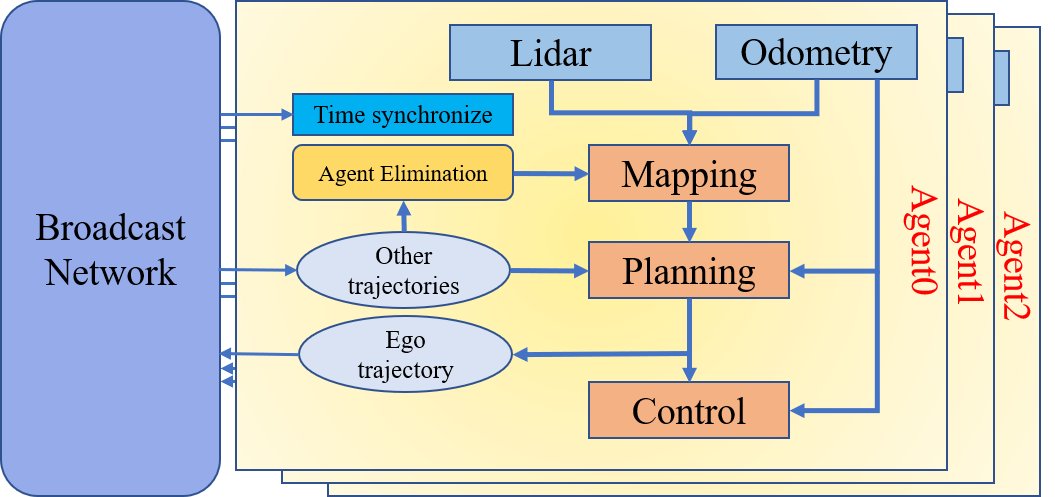}
	\setlength{\abovecaptionskip}{-0.4cm}
	\caption{System Architecture}
	\label{fig: system architecture}
	\vspace{-0.5cm}
\end{figure}
The overall system architecture is shown in Fig.\ref{fig: system architecture}. In this system, each agent is independent of each other, with trajectories broadcast by a wireless network. During navigation, each agent acquires point clouds and odometry information to construct 2-D grid maps. Then agents perform local planning based on the observed map and execute the trajectories by an MPC controller\cite{muske1993model}. 

The communication network is utilized in this system to broadcast trajectories and to synchronize timestamps among agents. 
As depicted in Eq.(\ref{equ: trajectory representation}), the information about trajectories only contains the coefficients and duration of the pieces, which means that the information size is small for modern wireless networks. Therefore, the communication delay can be ignored in our system.

The point cloud is the direct message from sensors such as lidars, which provides high-precision information about the obstacles. 
Since our front-end planning algorithm is based on spatial-temporal decomposition, the point clouds of dynamic obstacles would highly hinder the spatial path planning process, especially when the robots are traversing obstacle-dense environments. 
To eliminate the point clouds of other agents, we make use of the broadcast trajectories, and calculate dynamic agents' poses. Each agent is modeled as a convex polygon, and as a frame of point clouds is processed, each point is checked if it lies in the convex polygon. If so, the point is seen as the dynamic obstacle, and it is eliminated from the mapping process. By agent elimination, the remaining point clouds only come from static obstacles, freeing up enough space for path planning. 

Benefiting from the real-time performance, our system is capable of frequent replanning. 
While running, each agent conducts collision checks with static and dynamic obstacles. 
Replan is tackled as soon as a potential collision is detected, then a new trajectory is generated, simultaneously broadcast to other agents. 
In addition, replan is triggered if the agent has moved along half the length of the local trajectory, which aims to guarantee the motion coherence of the robot.

\begin{figure*}[t]
    \centering
    \includegraphics[width=17.55cm]{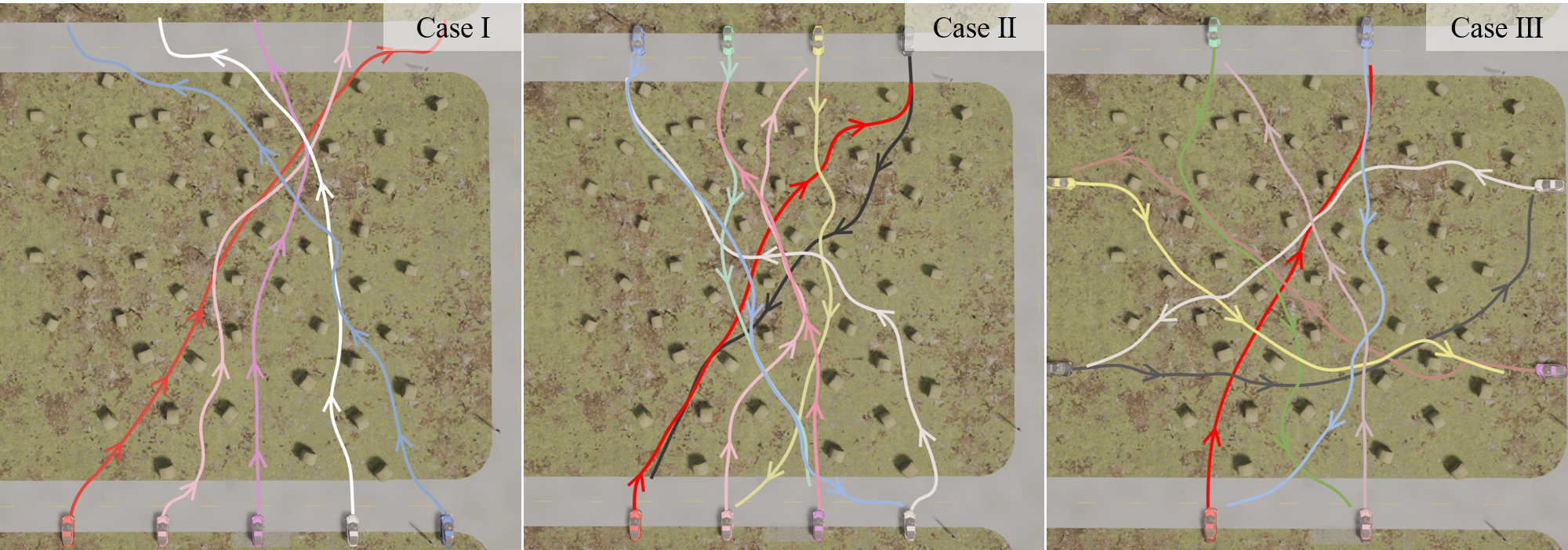}
    \setlength{\abovecaptionskip}{-0.0cm}
    \caption{Three cases in our simulation experiments.}
    \label{fig: three scenes combine}
    \vspace{-0.5cm}
\end{figure*}
\section{Experiments}
\label{sec: experiments}
We validate the robustness of our system in Carla\cite{Dosovitskiy17} and real-world experiments. In this section, we first introduce simulation experiments and ablation studies. Then real-world experiments are presented.
\subsection{Simulation Experiments} 
1) \textit{Environment Overview}: Simulation experiments are conducted on a farm scene, where obstacles are closely arranged. The size of the scene is $90m \times 90m$, where $48$ obstacles are randomly placed and there is no prior information about the environment, as shown in Fig. \ref{fig:HeadFigure}. We adopt Tesla Model 3 as the car model, where the maximum speed and acceleration are set as $8m/s$ and $3m/s^2$, respectively. 3D lidar is utilized as the mapping sensor, and the maximum sensing range is $30m$, which also serves as the horizon of the locally planned trajectory. For $SD$-VD parameters, we select $S=6m$, $D=1.5m$. CPU of the tested PC is Intel Core i7-10700, integrated with 16 GB RAM.

We simulate three cases: five cars driving from one side to the other, eight cars driving from two sides to the opposite sides, and eight cars driving from four sides to the opposite sides, respectively marked as case \uppercase\expandafter{\romannumeral1}, \uppercase\expandafter{\romannumeral2}, \uppercase\expandafter{\romannumeral3}, shown in Fig.\ref{fig: three scenes combine}. 

2) \textit{Computation Efficiency Evaluation}: In order to verify the efficiency of this system, we count the computation time spent in each module.
In addition, we compare the computation efficiency of the fixed time discretization strategy against our previous work to test the improvement.
The average computation time is counted in Tab.\ref{tab: time cost in planning}, where $t_{pp}$, $t_{sp}$, $t_{opt}$, and $t_{opt-ft}$ respectively mean the time spent in path planning, speed planning, spatial-temporal optimization and spatial-temporal optimization with the fixed time step discretization strategy. 
It is the fixed time step discretization strategy that decreases the computation time by nearly $20\%$. Besides, the average total time spent in planning is less than $30ms$ and the maximum total time is less than $100ms$, exhibiting strong real-time performance.

\begin{figure}[t]
    \vspace{-0.2cm}
    \centering
    \includegraphics[width=8.8cm]{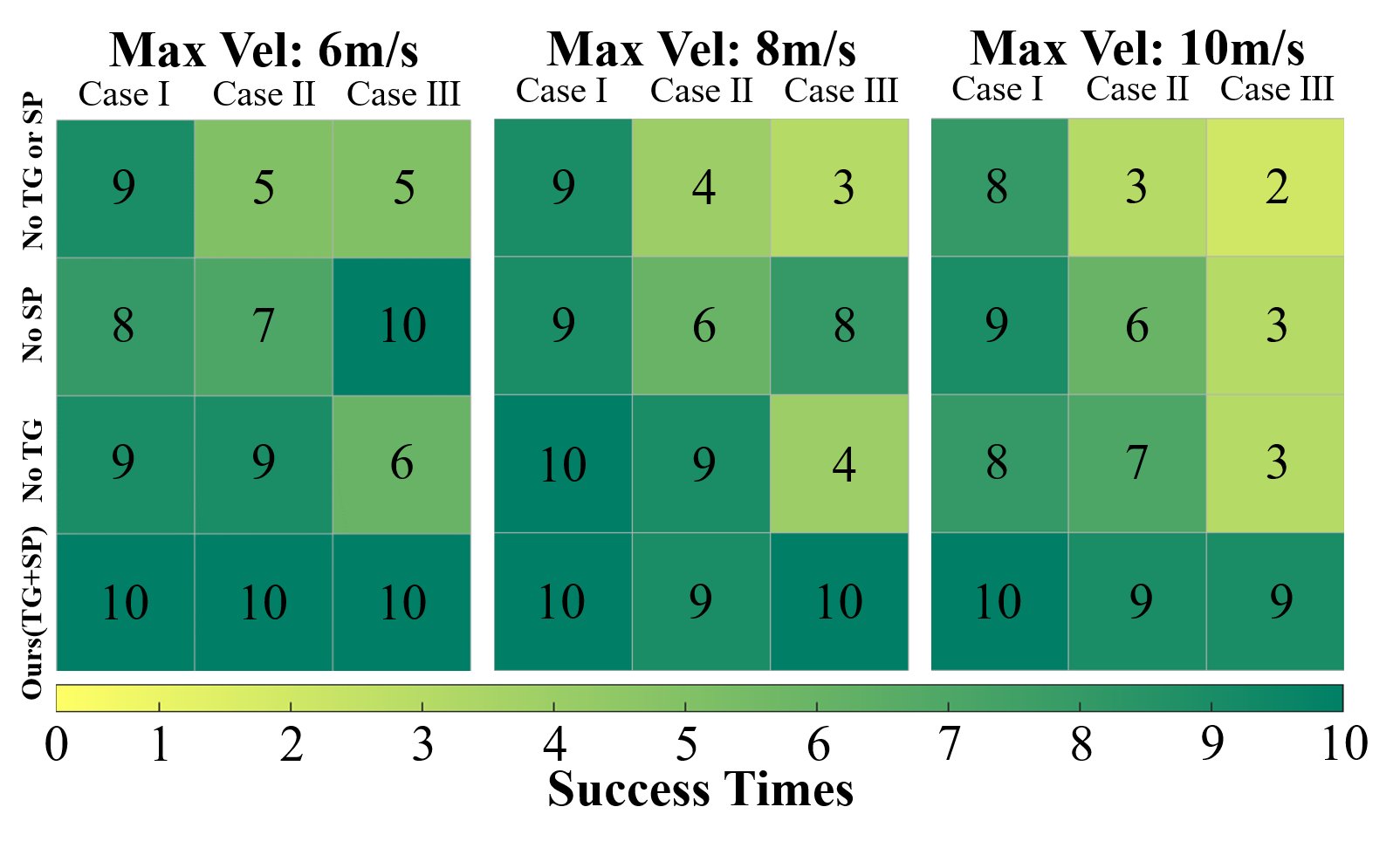}
    \setlength{\abovecaptionskip}{-0.8cm}
    \caption{Results of ablation studies in the simulation experiments.}
    \label{fig: color map}
    \vspace{-0.6cm}
\end{figure}

\begin{table}[t]
\centering
\caption{Time spent in planning}
\vspace{-0.25cm}
\begin{tabular}{c|cccc}
\toprule[1.5pt]
Items   & $t_{pp}$(ms) & $t_{sp}$(ms) & $t_{opt}$(ms) & $t_{opt-ft}$(ms)\\
\midrule[1pt]
Max     & 25.805         & 3.953         & 55.977   & 47.407       \\
Average & 7.399          & 0.295         & 21.123   & 16.259       \\
Min     & 2.431          & 0.078         & 3.449     & 2.714      \\ \hline
\end{tabular}
\label{tab: time cost in planning}
\vspace{-0.6cm}
\end{table}
3) \textit{Ablation Studies}: To evaluate the necessity of topology-guided path planning and search-based speed planning, we conduct ablation studies in the three cases above. 
In each case varying the maximum velocity, we perform $10$ trials, and the key metric is the number of successes that are collision-free. For the topology-guided path finding ablation, we eliminate the $SD$-VD class to cancel the trajectory consistency. 
For the search-based speed planning ablation, we directly adopt bang-bang control to provide an initial time arrangement. 
Results can be seen in Fig.\ref{fig: color map}, where TG and SP respectively represent Topological-Guidance and Speed-Planning. 
The results apparently show that the success rate falls off a cliff without TG or SP. 
Without topology-guided path planning, robot's trajectories lose homotopy under frequent replans, and agents cannot decide which side to move, luring the robots to collide with obstacles. 
Without speed planning, the initial values lead to a bad solution's local minima, raising a higher rate of colliding with other agents. 
In conclusion, the adopted topology-guided path planning and search-based speed planning effectively guarantees the safety of the robots.

\subsection{Real-world Experiments}
We conduct real-world experiments in a $3m\times8m$ environment, where three car-like robots move from one side to the other. 
Each robot is equipped with an NVIDIA Jetson Nano as the onboard processor and a single-line lidar with a $3m$ sensing range, which is equal to the horizon of locally planned trajectories.
Localization is done by the NOKOV motion capture system\footnote{https://www.nokov.com/}. 
WI-FI module is utilized to broadcast trajectories.
There is no prior map of the environment for the robots.
Besides, mapping, planning and control are all performed onboard. 
Maximum speed and acceleration are set as $0.6m/s$ and $1.0m/s^2$, respectively. The experiments are shown in Fig.\ref{fig:real world experiments}. 
Three agents manage to traverse the obstacle-rich environment, during which the obstacles are moved randomly, demonstrating robustness against the dynamic environment. 
For more information about the experiments, please watch our attached video.

\section{Conclusion}
In this paper, a decentralized framework for car-like robotic swarm in cluttered environments is proposed.
Topology-guided path planning and search-based speed planning are adopted to provide feasible initial values for the back-end optimization.
Afterward, spatial-temporal optimization is employed to generate a safe and smooth trajectory.
In each iteration of the optimization, the trajectory is discretized by fixed time steps, which reduces the computation burden.
Experiments demonstrate its high efficiency in real-time planning and low collision risk. 
However, there still exist some problems resulting from the deadlock, which also accounts for the occasional collision in our experiments.
In the future, we will try to solve these problems by introducing partial coordination and group planning.
After that, this framework will be extended to 2.5D environments.
% However, there still exists some problems such as no solution caused by spatial-temporal decomposition, which accounts for occasional collision in our experiments. 
% For instance, if two robots are moving in the opposite direction in a narrow space, then there is no solution by spatial-temporal decomposition, causing collision.
% In the future, we will focus on solving these problems by partial coordination and extend our swarm to broader scenarios.

% \bibliographystyle{ieeetr}
\bibliography{references}

\begin{thebibliography}{10}
\providecommand{\url}[1]{#1}
\csname url@rmstyle\endcsname
\providecommand{\newblock}{\relax}
\providecommand{\bibinfo}[2]{#2}
\providecommand\BIBentrySTDinterwordspacing{\spaceskip=0pt\relax}
\providecommand\BIBentryALTinterwordstretchfactor{4}
\providecommand\BIBentryALTinterwordspacing{\spaceskip=\fontdimen2\font plus
\BIBentryALTinterwordstretchfactor\fontdimen3\font minus
  \fontdimen4\font\relax}
\providecommand\BIBforeignlanguage[2]{{%
\expandafter\ifx\csname l@#1\endcsname\relax
\typeout{** WARNING: IEEEtran.bst: No hyphenation pattern has been}%
\typeout{** loaded for the language `#1'. Using the pattern for}%
\typeout{** the default language instead.}%
\else
\language=\csname l@#1\endcsname
\fi
#2}}

\bibitem{han2022differential}
Z.~Han, Y.~Wu, T.~Li, L.~Zhang, L.~Pei, L.~Xu, C.~Li, C.~Ma, C.~Xu, S.~Shen,
  \emph{et~al.}, ``Differential flatness-based trajectory planning for
  autonomous vehicles,'' \emph{arXiv preprint arXiv:2208.13160}, 2022.

\bibitem{Dosovitskiy17}
A.~Dosovitskiy, G.~Ros, F.~Codevilla, A.~Lopez, and V.~Koltun, ``{CARLA}: {An}
  open urban driving simulator,'' in \emph{Proceedings of the 1st Annual
  Conference on Robot Learning}, 2017, pp. 1--16.

\bibitem{zhu2015convex}
Z.~Zhu, E.~Schmerling, and M.~Pavone, ``A convex optimization approach to
  smooth trajectories for motion planning with car-like robots,'' in \emph{2015
  54th IEEE conference on decision and control (CDC)}.\hskip 1em plus 0.5em
  minus 0.4em\relax IEEE, 2015, pp. 835--842.

\bibitem{zhou2020autonomous}
J.~Zhou, R.~He, Y.~Wang, S.~Jiang, Z.~Zhu, J.~Hu, J.~Miao, and Q.~Luo,
  ``Autonomous driving trajectory optimization with dual-loop iterative
  anchoring path smoothing and piecewise-jerk speed optimization,'' \emph{IEEE
  Robotics and Automation Letters}, vol.~6, no.~2, pp. 439--446, 2020.

\bibitem{zhang2020optimization}
X.~Zhang, A.~Liniger, and F.~Borrelli, ``Optimization-based collision
  avoidance,'' \emph{IEEE Transactions on Control Systems Technology}, vol.~29,
  no.~3, pp. 972--983, 2020.

\bibitem{li2021optimization}
B.~Li, T.~Acarman, Y.~Zhang, Y.~Ouyang, C.~Yaman, Q.~Kong, X.~Zhong, and
  X.~Peng, ``Optimization-based trajectory planning for autonomous parking with
  irregularly placed obstacles: A lightweight iterative framework,'' \emph{IEEE
  Transactions on Intelligent Transportation Systems}, 2021.

\bibitem{alonso2018cooperative}
J.~Alonso-Mora, P.~Beardsley, and R.~Siegwart, ``Cooperative collision
  avoidance for nonholonomic robots,'' \emph{IEEE Transactions on Robotics},
  vol.~34, no.~2, pp. 404--420, 2018.

\bibitem{alonso2012reciprocal}
J.~Alonso-Mora, A.~Breitenmoser, P.~Beardsley, and R.~Siegwart, ``Reciprocal
  collision avoidance for multiple car-like robots,'' in \emph{2012 IEEE
  International Conference on Robotics and Automation}.\hskip 1em plus 0.5em
  minus 0.4em\relax IEEE, 2012, pp. 360--366.

\bibitem{9812126}
Y.~Ouyang, B.~Li, Y.~Zhang, T.~Acarman, Y.~Guo, and T.~Zhang, ``Fast and
  optimal trajectory planning for multiple vehicles in a nonconvex and
  cluttered environment: Benchmarks, methodology, and experiments,'' in
  \emph{2022 International Conference on Robotics and Automation (ICRA)}, 2022,
  pp. 10\,746--10\,752.

\bibitem{9345421}
B.~Li, Y.~Ouyang, Y.~Zhang, T.~Acarman, Q.~Kong, and Z.~Shao, ``Optimal
  cooperative maneuver planning for multiple nonholonomic robots in a tiny
  environment via adaptive-scaling constrained optimization,'' \emph{IEEE
  Robotics and Automation Letters}, vol.~6, no.~2, pp. 1511--1518, 2021.

\bibitem{8258896}
I.~M. Delimpaltadakis, C.~P. Bechlioulis, and K.~J. Kyriakopoulos,
  ``Decentralized platooning with obstacle avoidance for car-like vehicles with
  limited sensing,'' \emph{IEEE Robotics and Automation Letters}, vol.~3,
  no.~2, pp. 835--840, 2018.

\bibitem{jaillet2008path}
L.~Jaillet and T.~Sim{\'e}on, ``Path deformation roadmaps: Compact graphs with
  useful cycles for motion planning,'' \emph{The International Journal of
  Robotics Research}, vol.~27, no. 11-12, pp. 1175--1188, 2008.

\bibitem{zhou2020robust}
B.~Zhou, F.~Gao, J.~Pan, and S.~Shen, ``Robust real-time uav replanning using
  guided gradient-based optimization and topological paths,'' in \emph{2020
  IEEE International Conference on Robotics and Automation (ICRA)}.\hskip 1em
  plus 0.5em minus 0.4em\relax IEEE, 2020, pp. 1208--1214.

\bibitem{zhou2021raptor}
B.~Zhou, J.~Pan, F.~Gao, and S.~Shen, ``Raptor: Robust and perception-aware
  trajectory replanning for quadrotor fast flight,'' \emph{IEEE Transactions on
  Robotics}, vol.~37, no.~6, pp. 1992--2009, 2021.

\bibitem{zhou2021ego}
X.~Zhou, J.~Zhu, H.~Zhou, C.~Xu, and F.~Gao, ``Ego-swarm: A fully autonomous
  and decentralized quadrotor swarm system in cluttered environments,'' in
  \emph{2021 IEEE International Conference on Robotics and Automation
  (ICRA)}.\hskip 1em plus 0.5em minus 0.4em\relax IEEE, 2021, pp. 4101--4107.

\bibitem{cheng2022real}
J.~Cheng, Y.~Chen, Q.~Zhang, L.~Gan, C.~Liu, and M.~Liu, ``Real-time trajectory
  planning for autonomous driving with gaussian process and incremental
  refinement,'' in \emph{2022 International Conference on Robotics and
  Automation (ICRA)}.\hskip 1em plus 0.5em minus 0.4em\relax IEEE, 2022, pp.
  8999--9005.

\bibitem{fan2018baidu}
H.~Fan, F.~Zhu, C.~Liu, L.~Zhang, L.~Zhuang, D.~Li, W.~Zhu, J.~Hu, H.~Li, and
  Q.~Kong, ``Baidu apollo em motion planner,'' \emph{arXiv preprint
  arXiv:1807.08048}, 2018.

\bibitem{liu2017speed}
C.~Liu, W.~Zhan, and M.~Tomizuka, ``Speed profile planning in dynamic
  environments via temporal optimization,'' in \emph{2017 IEEE Intelligent
  Vehicles Symposium (IV)}.\hskip 1em plus 0.5em minus 0.4em\relax IEEE, 2017,
  pp. 154--159.

\bibitem{xu2022speed}
W.~Xu and J.~M. Dolan, ``Speed planning in dynamic environments over a fixed
  path for autonomous vehicles,'' in \emph{2022 International Conference on
  Robotics and Automation (ICRA)}.\hskip 1em plus 0.5em minus 0.4em\relax IEEE,
  2022, pp. 3321--3327.

\bibitem{johnson2012optimal}
J.~Johnson and K.~Hauser, ``Optimal acceleration-bounded trajectory planning in
  dynamic environments along a specified path,'' in \emph{2012 IEEE
  International Conference on Robotics and Automation}.\hskip 1em plus 0.5em
  minus 0.4em\relax IEEE, 2012, pp. 2035--2041.

\bibitem{johnson2013optimal}
------, ``Optimal longitudinal control planning with moving obstacles,'' in
  \emph{2013 IEEE Intelligent Vehicles Symposium (IV)}.\hskip 1em plus 0.5em
  minus 0.4em\relax IEEE, 2013, pp. 605--611.

\bibitem{li2021speed}
J.~Li, X.~Xie, H.~Ma, X.~Liu, and J.~He, ``Speed planning using bezier
  polynomials with trapezoidal corridors,'' \emph{arXiv preprint
  arXiv:2104.11655}, 2021.

\bibitem{gonzalez2016speed}
D.~Gonz{\'a}lez, V.~Milan{\'e}s, J.~P{\'e}rez, and F.~Nashashibi, ``Speed
  profile generation based on quintic b{\'e}zier curves for enhanced passenger
  comfort,'' in \emph{2016 IEEE 19th international conference on intelligent
  transportation systems (ITSC)}.\hskip 1em plus 0.5em minus 0.4em\relax IEEE,
  2016, pp. 814--819.

\bibitem{dolgov2010path}
D.~Dolgov, S.~Thrun, M.~Montemerlo, and J.~Diebel, ``Path planning for
  autonomous vehicles in unknown semi-structured environments,'' \emph{The
  international journal of robotics research}, vol.~29, no.~5, pp. 485--501,
  2010.

\bibitem{doi:10.1137/1022026}
\BIBentryALTinterwordspacing
Z.~Artstein, ``Discrete and continuous bang-bang and facial spaces or: Look for
  the extreme points,'' \emph{SIAM Review}, vol.~22, no.~2, pp. 172--185, 1980.
  [Online]. Available: \url{https://doi.org/10.1137/1022026}
\BIBentrySTDinterwordspacing

\bibitem{fuchshumer2005nonlinear}
S.~Fuchshumer, K.~Schlacher, and T.~Rittenschober, ``Nonlinear vehicle dynamics
  control-a flatness based approach,'' in \emph{Proceedings of the 44th IEEE
  Conference on Decision and Control}.\hskip 1em plus 0.5em minus 0.4em\relax
  IEEE, 2005, pp. 6492--6497.

\bibitem{wang2022geometrically}
Z.~Wang, X.~Zhou, C.~Xu, and F.~Gao, ``Geometrically constrained trajectory
  optimization for multicopters,'' \emph{IEEE Transactions on Robotics}, 2022.

\bibitem{muske1993model}
K.~R. Muske and J.~B. Rawlings, ``Model predictive control with linear
  models,'' \emph{AIChE Journal}, vol.~39, no.~2, pp. 262--287, 1993.

\end{thebibliography}

\end{document}